\documentclass[lettersize,journal]{IEEEtran}
\usepackage{amsmath,amsfonts}
\usepackage{algorithmic}
\usepackage{algorithm}
\usepackage{array}
\usepackage[caption=false,font=normalsize,labelfont=sf,textfont=sf]{subfig}
\usepackage{textcomp}
\usepackage{stfloats}
\usepackage{url}
\usepackage{verbatim}
\usepackage{graphicx}
\usepackage{cite}
\usepackage{float}
\usepackage{xcolor}
\usepackage{amsfonts}
\usepackage{booktabs}
\usepackage{siunitx}
\usepackage{array}
\floatstyle{plaintop}
\restylefloat{table}
\usepackage{tabularx}
\usepackage{multirow}
\newcommand\mcl[1]{\multicolumn{2}{|l|}{#1}}
\begin{document}

\title{IMUNet: Efficient Regression Architecture for IMU Navigation and Positioning}

\author{\IEEEauthorblockN{Behnam Zeinali\IEEEauthorrefmark{1},
        Hadi Zandizari\IEEEauthorrefmark{2}, 
        and~J. Morris Chang\IEEEauthorrefmark{3} \\}
\IEEEauthorblockA{Department of Electrical Engineering \\
University of South Florida \\ Tampa, Florida 33620\\
Email: \IEEEauthorrefmark{1}behnamz@usf.edu, 
\IEEEauthorrefmark{2}hadiz@usf.edu,
\IEEEauthorrefmark{3}chang5@usf.edu}

\thanks{
}
\thanks{Manuscript received July 4, 2022.
All the source codes are available at: https://github.com/BehnamZeinali/IMUNet.
}}

\markboth{Submitted to IEEE Transactions on Mobile Computing
}%
{Shell \MakeLowercase{\textit{et al.}}: A Sample Article Using IEEEtran.cls for IEEE Journals}


\maketitle

\begin{abstract}

Data-driven based method for navigation and positioning has absorbed attention in recent years and it outperforms all its competitor methods in terms of accuracy and efficiency. This paper introduces a new architecture called IMUNet which is accurate and efficient for position estimation on edge device implementation receiving a sequence of raw IMU measurements. The architecture has been compared with one dimension version of the state-of-the-art CNN networks that have been introduced recently for edge device implementation in terms of accuracy and efficiency. Moreover, a new method for collecting a dataset using IMU sensors on cell phones and Google ARCore API has been proposed and a publicly available dataset has been recorded. A comprehensive evaluation using four different datasets as well as the proposed dataset and real device implementation has been done to prove the performance of the architecture. All the code in both Pytorch and Tensorflow framework as well as the Android application code have been shared to improve further research.

\end{abstract}


\section{Introduction}

\IEEEPARstart{T}{}hree-axis gyroscope and a three-axis accelerometer make up a common inertial measurement unit (IMU). These sensors 
assess a moving platform's rotational velocity and linear acceleration. Some IMU sensors provide the gravity-compensated linear acceleration of a moving platform as well. Since the IMU collects motion data without the use of any external infrastructure, it's been utilized in a variety of navigation systems and applications, including edge device applications \cite{li2013high}, \cite{9134860}, robotics \cite{zhang2021pose}, drones \cite{nisar2019vimo}, automatic vehicles \cite{levinson2011towards}, and so on. 

Most methods for processing IMU data rely on hand-crafted models to approximate sensor properties and the dynamic characteristics of the underlying motion. In \cite{10.5555/1594745} the pipeline for processing the IMU data has been summarized as follows: filtering the data, model to compensate for the intrinsic characteristics of the sensors, data estimation using interpolation and extrapolation techniques, integration techniques for pose calculation, and sensor fusion techniques to compensate measurement drifting. Filtering the data and model compensation are called the preprocessing steps which mainly be used to obtain cleaned data by eliminating the error caused by the intrinsic properties of the sensors or collected during the measuring process. 

Interpolation, extrapolation, and integration techniques are common methods to determine the IMU's integrated orientation and position based on the underlying motion dynamics. Since there is an inevitable drifting during the IMU integration,  data from additional sensors \cite{li2013high}, \cite{nisar2019vimo}, or motion restrictions \cite{Abdulrahim2014UnderstandingTP}, \cite{wagstaff2018lstm} is also harnessed to update sensors' data using probabilistic method to mitigate the measurement drift.

\begin{figure}
  \centering
  \includegraphics[width=3.8in]{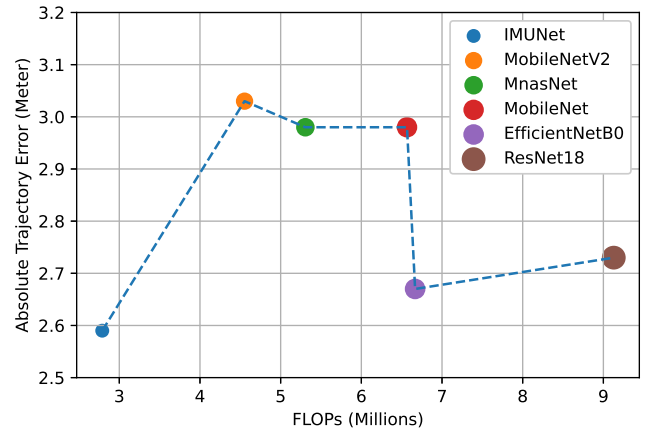}
  \caption{The number of FLOPs and the accuracy of the
  proposed dataset for each architecture. (The number of FLOPs is for one dimension version of the architectures)}
  \label{fig:proposed_flops}
\end{figure}

The current IMU data preprocessing methods are appropriate to produce noise-free data for further processing. However, they are still susceptible to some error resources. The sensor modeling error is one of the most significant steps in IMU data pre-processing. In \cite{Abdulrahim2014UnderstandingTP} and \cite{HGJHC0_2011_v12n4_371}, authors studied the signal processing model for IMU sensors, while in \cite{trawny2005indirect} and \cite{schneider2017visual}, the intrinsic model for sensor modeling error has been investigated. However, these models are handcrafted and being designed by considering engineering limitations and cannot be considered as a general solution across all platforms and applications.

In this paper, we offer a new machine learning architecture for a data-driven strategy for inertial sensor modeling to achieve improved performance and efficiency inspired by recent papers harnessing data-driven methods and neural networks for positioning, navigation, and timing \cite{wagstaff2018lstm}, \cite{chen2018ionet, jiang2018mems, ronin, brossard2020denoising, zhang2021imu}. An edge-device-friendly deep neural network has been designed in this paper which processes and fuses the IMU data measurements and regresses the position and the velocity as an output. Additionally, a method for collecting a dataset using an Android mobile phone has been proposed in this paper and a new dataset has been collected. The method for collecting the dataset is the modification of the method in \cite{cortes2018advio} and it is similar to the method that has been introduced in \cite{ridi}. However, instead of using the Tango device, our proposed method is collecting the ground truth data using Google ARCore API \cite{arcore} which is introduced by Google in 2019. By doing so, one will be able to collect the dataset using recent android phones.
\begin{figure*}
  \centering
    \includegraphics[width=\linewidth]{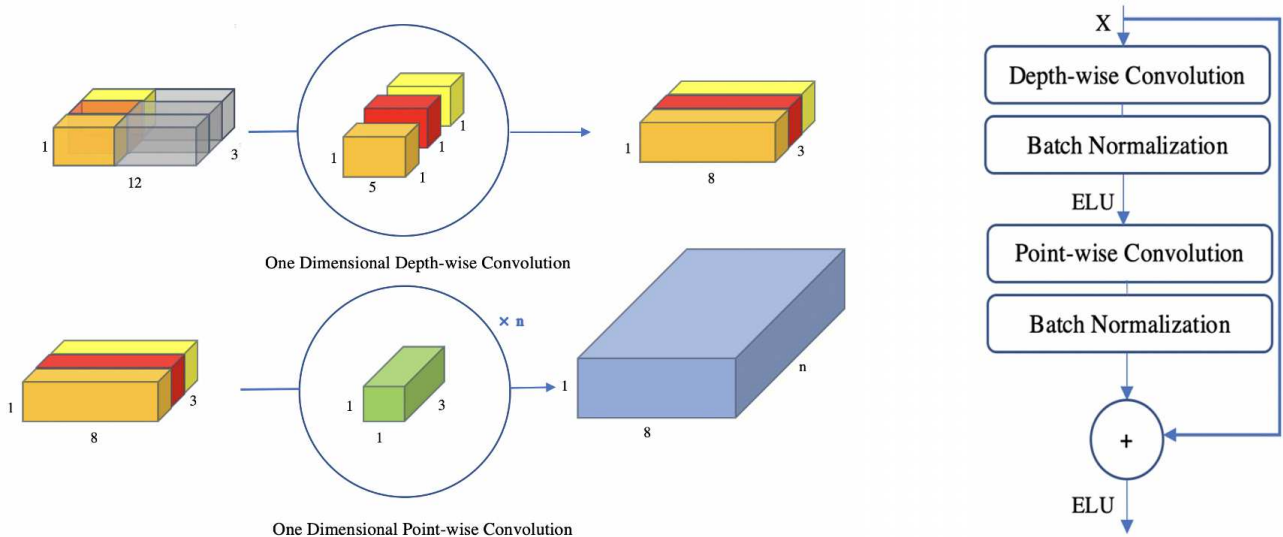}
  \caption{Left: Depth-wise and Point-wise convolution for one dimensional data. Right: The Proposed MobileResnet Block }
  \label{fig:proposed_imunet}
\end{figure*}
Other than the proposed architecture, the performance of some of the state-of-the-art edge device-friendly CNN architectures such as MobileNet\cite{mobilenets}, MobileNetV2\cite{mobilenetv2}, MnasNet\cite{mnasnet}, and EfficientB0\cite{efficientnet} have been investigated while using them as a regression model for data-driven methods. All the architectures have been modified to be compatible with IMU measurements. Consequently, these models have been used as a regression model for the recent successful data-driven method \cite{ronin} and the performance of them, the proposed model, and the Resnet18 architecture \cite{resnet} which is the model that has been used in \cite{ronin} has been evaluated and reported on quite a few datasets.

To summarize, our contributions are three-fold:
\begin{itemize}
 \item A new architecture called IMUNet which is appropriate for edge device implementation using IMU measurements is proposed to do the inertial navigation efficiently and accurately while preserving more energy.
  \item An improved version of the method used in\cite{cortes2018advio} by modifying the method introduced in\cite{ridi} for collecting a dataset using the Google ARCore API has been proposed and a publicly available dataset has been collected.
 
  \item An empirical study has been implemented for evaluating the performance of the proposed model as well as different state-of-the-art CNN models on different datasets.
  \item Tensorflow and Pytorch versions of the framework are implemented. Also, the testing part of the data-driven methods has been implemented for Android devices and all the source codes, as well as the android application code for the testing part and collecting a new dataset, are publicly available.  
\end{itemize}

The rest of this paper is organized as follows:
In section ~\ref{sec:related_work} we will scrutinize different efforts and strategies that have been done so far by researchers to do inertial navigation and positioning using IMU sensors. In section~\ref{sec:proposed_dataset} we will introduce the proposed method for collecting the dataset in depth. In section~\ref{sec:proposed_model}, we will walk through the preliminary knowledge about inertial navigation using double integration methods, data-driven strategy, and the proposed architecture. The experimental result will be thoroughly investigated in section~\ref{sec:experiment_result}. Section~\ref{sec:conclusion} concludes the paper.

\section{Related work}\label{sec:related_work}

IMU sensors are susceptible to a variety of error sources. These errors are caused by electrical, mechanical, and flaws in signal processing systems. Statistical methods are the most available strategies that are being used to minimize the noise in IMU reading data and adjust the error in the intrinsic model. The high-frequency components are typically eliminated or attenuated when filtering IMU measurement output in inertial navigation methods \cite{HGJHC0_2011_v12n4_371}. In \cite{s130809549}, the authors used Auto regressing and moving average (ARMA) methods for this end.  The additional components can be considered in measurement's equations to handle the IMU sensors biases which affect IMU readings other than noises \cite{trawny2005indirect}.

Although handling the noise and sensors biased can be sufficient to have an acceptable accuracy when dealing with high-end IMU sensors when dealing with low-cost IMU sensors, other parameters must be taken into accounts such as G-sensitivity matrix and misalignment and scale factors \cite{li2014high, Yang2020OnlineII}. In \cite{niu2013fast}, the authors considered the thermal components that can affect the final accuracy to have a better real application. Using Newton’s law \cite{li2013high, 10.5555/1594745} and numerical integration, the inertial position can be calculated when the cleaned and error-free measurement is available. 
Mathematical descriptions of the sensor's noises and the intrinsic model method are incapable of attaining clean data since the error is highly dependent on the type of IMU sensors, signal processing approaches, and measurement procedures.

Even with an appropriate mathematical description of the noise, drifting will be accrued in the pose integration. To overcome the accumulated drifting, one approach is to fuse the measurements from other sensors such as, visual sensors \cite{li2013high}, a global positioning system (GPS) \cite{10.5555/1594745}, wheel odometers \cite{zhang2021pose}, and laser range finders \cite{geneva2018lips} with the IMU reading. Another approach is to apply constraints to the motion or consider a repetitive motion pattern in the platform to alleviate the estimation error by using the probability techniques \cite{Abdulrahim2014UnderstandingTP, wagstaff2018lstm, ahmed2018visual}. Zero-velocity event \cite{Abdulrahim2014UnderstandingTP, wagstaff2018lstm} is the typical method to apply the constraints to the motion pattern. In these methods, IMU is mounted on the foot and the drifting is corrected considering the constraint that when the foot touched the ground, the velocity is zero. However, these methods require an additional foot-mounted sensor when IMU sensors mounted on edge devices are supposed to be used. Moreover, for drone navigation, these methods are useless.

Pedestrian pattern methods \cite{ahmed2018visual} are the methods that consider a repetitive motion pattern in the platform. For those motions that are inherently repetitive like human motions, heuristics of the motion can be harnessed to estimate the navigation. Step counting is a method that considers some conditions to estimate tracking and navigation. It considers that the tracking distance is related to the number of steps and the IMU sensors are providing data rigidly during the tracking. Considering this controlled environment, the results have been acceptable and impressive. In \cite{6851371}, the authors tried to use a more elaborated method using frequency domain analysis to estimate the motion direction. Some other researchers have tried to take advantage of both considering the controlled motion and using additional sensors. In \cite{6696807}, by taking the integration from the hovering motion detection module, Kottas et al. proposed a visual-inertial odometry method. In \cite{nisar2019vimo}, VIMO is proposed which is a method that estimates the sensor poses using multi-sesnors' inputs. Although the performance of these methods is acceptable, they do not show the robustness of the data-driven-based methods.v

Recently, data-driven strategy is another method that is widely being investigated to overcome the accumulated drifting during the integration.  Deep Neural Networks(DNNs) have been showing acceptable performances in a variety of different applications it has inspired researchers to increase the inertial navigation accuracy using these methods to constitute IMU models \cite{9134860, wagstaff2018lstm}. These models are responsible to take the end-to-end integration on the IMU sensors data and estimate the IMU poses \cite{ridi, ronin,zhang2021imu}. These models are also harnessed to represent sensor models \cite{jiang2018mems,brossard2020denoising}. In \cite{deep_nav}, a recurrent neural network model has been used to estimate the position as a regression model. These models are more appropriate for series data and transfer memory knowledge from the previous stage to the later points. IMU, Magnetometer, and barometer measurements have been used as inputs and the output is the position and the velocity. However, it is not feasible to use these methods for general cases since the network parameters need to be trained for each IMU sensor output specifically.

\section{Proposed dataset}\label{sec:proposed_dataset}

A variety of methods have been proposed so far to collect the datasets. Different tools have been harnessed to collect the IMU sensors data such as using the IMU sensors on edge devices or developing an embedded system with IMU sensors on the chip. The former one mostly used for inertial navigation inside and outside the buildings and the latter is being used for vehicle and drone tracking in the wild. One of the most challenging parts while collecting the dataset for inertial navigation is to obtain the accurate and precise trajectory of the navigation as ground truth to evaluate the different methods. Using additional sensors other than IMU measurements is the most common method that has been used for this purpose. In the following, some of the recent publicly available datasets have been presented.

\subsection{Available Datasets} \label{sec:available_dataset}
\subsubsection{PX4 dataset}\label{sec:px4_dataset}
In \cite{5980229}, an inexpensive flight controller board called Pixhawx which is appropriate for research as well as industrial purposes has been used to collect the dataset. The board integrates IMU sensors, magnetometer, flight controlling, monitoring components, and an ARM processor as well as a micro SD card for recording the requiring IMU sensors data and flight information. An open-source autopilot software named PX4 autopilot has been developed by people who developed the first Pixhawk flight controller \cite{7140074}. The software features control loops, planning algorithm, and sensor drivers as well as a library named Estimation and Control Library (ECL) \cite{paul_riseborough} with a sensor fusion method called EKF2 \cite{GARCIA2020136}. EKF2 is a sensor fusion algorithm that fuses the GPS, Magnetometer, Barometer, and IMU sensors measurements to estimate the actual trajectory, velocity, and quaternion representation of the attitude in the board's frame (local North-East-Down frame). This information can be used as ground truth for data-driven method purposes by researchers. A database contains the flight logs of real flight data that are being collected and uploaded by users all around the world. The logs are being maintained by the PX4 team. Thousands of flight logs of different vehicles and various Pixhawk versions can be found in the database \cite{px4}.

\subsubsection{RIDI dataset}\label{sec:ridi_dataset}
In \cite{ridi}, a dataset called RIDI was collected using IMU sensors in smartphones. The actual position of the phone has been calculated using the Visual Inertial Odometery system \cite{project_tango} as a ground truth. A Lenovo Phab2 Pro smartphone which is armed with Google Tango augmented reality features has been used to collect the angular velocities, Magnetometer, linear acceleration measurements as well as 3D poses of the camera and the orientation of the device using the SLAM (Simultaneous Localization and Mapping)\cite{slam} method. Android APIs have been used to obtain the device orientation and the inertial odometry system on Tango has been used to calculate the camera's poses. 100 minutes of walking from six different people and four different placements of the phone have been collected. The trajectories contain a variety of motions such as side motion, backward, forward walking, and acceleration/deceleration motions. The data has been collected at 200Hz which is the camera frame rate of the Lenovo phone and other sensors' measurements have been synchronized into the time-stamps of the camera using the linear interpolation methods. An android application for collecting the dataset has been published by the author, however, one must have a device with Tango features to be able to use the application. Another drawback of this dataset is that the camera must have a clear field of view all the time and only the measurements of the IMU sensors integrated on the Lenovo device have been collected which limits the generalization of the method for real-time implementation purposes.

\subsubsection{OXIOD dataset}\label{sec:oxiod_dataset}
In \cite{oxiod}, a large dataset named OXIOD has been proposed which uses a motion capture system called (Vicon) which precisely captures the motions. Various phone placements such as a hand, a bag, a pocket, and a trolley have been harnessed for data collection. Five human subjects collected the dataset and in total 14.7 hours of data have been collected. Similar to the previous dataset, a single device has been used for both ground truth and IMU sensor measurements collections, and the camera must have a clear sight of view for the Visual Inertial SLAM method or should be clearly visible for the Vicon system all the time which makes the method unusable for natural phone positionings such as in a bag or leg pocket.   
\subsubsection{RONIN dataset}\label{sec:ronin_dataset}
In \cite{ronin}, the authors enhanced their previous data collection methods\cite{ridi} to address its mentioned drawbacks. A two-device data gathering method has been proposed while one phone has been used to collect the IMU measurements as well as the magnetometer and the barometer measurements in a natural way such as day-to-day activities so that the phone can be held in the pocket or in the bag during different movements such as sitting and walking around. For collecting the ground truth, a similar SLAM method has been harnessed using a 3D tracking phone (Asus Zenfone AR) which is attached to the subject's body so that instead of the trajectory of the phone, the trajectory of the body can be estimated using data-driven methods. A data preprocessing pipeline is harnessed to align two different phones and overcome the drifting. 42.7 ours of IMU measurements data from 100 human subjects and two different android phones inside three different buildings have been collected while users can freely hold the second phone like a natural phone handling such as putting it deep in the bag or packet. Both phones collected the data at 200Hz. However, to collect the ground truth, a specific 3D tracking phone is required which limits the data collection.

\subsection{Proposed data collection method}
All the last three mentioned suffering from ground truth data collection since they are required a specific type of edge device to collect the actual trajectory of the phone. This drawback limits their general usage for people. One of the contributions of this article is to address this problem. Our data collection is similar to the method that has been used in \cite{ridi}, However, instead of using a specific phone (i.e. Lenovo or Asus Zenfone), we have used ARCore API to harness the SLAM technique for the ground truth collection. Using our proposed method, any android phone whose camera features motion tracking ability (more than 90\% of these days phones) can be used to collect the ground truth data. The ARCore API allows you to collect the camera pose at 30Hz or 60Hz if the camera features 60Hz data collections. However, the sampling rate of the IMU sensors is higher than this number. By synchronizing the IMU sensors' data to the camera pose time-stamp, IMU sensors will be down-sampled, and losing these measurements will hurt the dataset. A prepossessing pipeline that uses linear interpolation to synchronize the camera time-stamp with any sampling rate target has been used to preserve the data. Then, another linear interpolation has been applied to all IMU sensor measurements to synchronize them with the new sampling rate time-stamp. Also, Google Tango phone, Lenovo
Phab2 Pro with a similar method in\cite{ridi} has been harnessed to collect the dataset. Consequently, two different datasets have been collected:

\begin{itemize}
  \item A dataset that uses the exact method in \cite{ridi} using Google Tango phone, Lenovo Phab2 Pro
  \item A dataset in which the ARCore API has been harnessed for ground truth collection using the Samsung S10 smartphone
 
\end{itemize}

For all the datasets, two human subjects collect the data by holding the phone by hand. About 60 minutes of data have been collected for each set of data. The android application is also provided that allows the majority of android device users to collect their own dataset in their own scenarios and conditions. These two datasets have been combined and used as a proposed dataset in this paper.

\section{IMU Proposed Methodology MODELS}\label{sec:proposed_model}
\subsection{Double Integration Method}
In a positioning system using the double integration method, taking integration twice from noise and drift-free IMU measurements in the global frame will provide the position. However, the measurements are in the IMU frame and noisy. Moreover,  drifting is inevitable after measuring the IMU signals. The IMU framework can be defined as follow:

\begin{equation}
\label{deqn_ex1a}
G_g, A_g = f(G_{imu} , A_{imu}, err_{m})
\end{equation}

Where $G_g$ and $A_g$ are the noise and drifting-free gyroscope and accelerometer measurements in the global frame. $G_{imu}$ and $A_{imu}$ are the noise and drifting-free gyroscope and accelerometer measurements in the imu frame and  $err_{m}$ is the measurement's error which contains the drifting and the noise. The framework $f$ plays an important role and can be divided into two main parts: noise cancellation and drifting prevention. For noise cancellation, the sensor modeling technique is being used in a non-linear relationship. Let's consider $W$ as a sensor modeling system. then applying noisy IMU measurements to this model would provide the noise-free measurements using this equation:
\begin{equation}
\label{noise_imu}
IMU_{c} = W*{IMU_{n}}
\end{equation}
Where $IMU_{c}$ and $IMU_{n}$ are clean and noisy IMU data respectively. For drifting prevention, a constraint will apply to the framework to compensate for the drifting.
\begin{equation}
\label{drift_imu}
IMU_{m} = IMU_{c}  + IMU_{d}
\end{equation}
Where $IMU_{d}$ is the drifting values and the constraint tries to estimate the drifting values and subtract them from the measurement to achieve the clean IMU signals.  The orientation can be obtained using the sensor fusion techniques using gyroscope, barometer, and magnetometer sensors. Using the quaternion multiplication, data from the IMU frame can be transformed into the global frame for further processing:
\begin{equation}
IMU_{g} = Ori**{IMU_{imu}}
\end{equation}
Where $**$ shows the quaternion multiplication, $Ori$ is the orientation vector, and $IMU_{g}$ and $IMU_{imu}$ are IMU measurements in the global and IMU frame respectively. While achieving the clean IMU measurements in the global frame, the Velocity in the global frame $V_g$ in the time frame $t$ can be calculated using the Newton law as follow:

\begin{equation}
\label{vel_integration}
V_g(t_{k+1}) = \sum_{t=t_k}^{t_{k+1}} A_g(t)*\Delta(t_{k+1}-{t_k}) + V_g(t_k) 
\end{equation}

By taking another integration from the velocities, the position values in the global frame can be obtained:
\begin{equation}
\label{pose_integration}
P_g(t_{k+1}) = \sum_{t=t_k}^{t_{k+1}} V_g(t)*\Delta(t_{k+1}-{t_k}) + P_g(t_k) 
\end{equation}

The data-driven methodology tries to perform integration, sensor modeling, and drifting prevention using a powerful machine learning architecture. The details of the restrictions, and the learning method. and cost functions can be different, however, the strategy is the same. In \cite{ronin}, the raw IMU measurements are being applied to a model to estimate the clean velocity data:
\begin{equation}
\label{ml_eq}
V_g(t_{k+1}) = ML_p(Inputs) 
\end{equation}
Where $ML$ is a regression machine learning model and $V_g(t_{k+1})$ is the velocity in the global frame. $P$ is the parameters of the network and should be trained using the back propagation method to do equations~\ref{noise_imu},~\ref{drift_imu}, and~\ref{vel_integration} simultaneously. Achieving the estimated values for $V_g(t_{k+1})$, the position can be obtained using the equation~\ref{pose_integration}. The $Inputs$ in equation~\ref{ml_eq} can be obtained from the equation below:
\begin{equation}
Inputs = Ori**(A_{imu}(t_{k->k+1}), G_{imu}(t_{k->k+1}))
\end{equation}
Where $Ori$ is the orientation, $**$ is the quaternion multiplication, $A_{imu}$ and $G_{imu}$ are raw accelerometer and gyroscope measurements in the IMU frame respectively. $t_{k->k+1}$ shows all the measurements from $t_k$ to $t_{k+1}$. For training the model, the ground truth velocity can be measured using the sensors or by taking a derivative from the ground truth position obtained by external sensors such as a camera or GPS sensors.

\subsection{Data-driven methods}\label{sec:data-driven methods}
Data-driven methods have absorbed quite a little attention recently for inertial navigation purposes. In these methods, a machine learning model is being used to provide noise-free IMU values or even estimate the velocity/position as an output with receiving the raw noisy IMU values as an input. For doing so, the ground truth velocity/position value is required for training the model. These values can be obtained from other sensors such as GPS or the camera. A data-driven method introduced in \cite{ronin} has been used as a based method in this paper. The author proposed a method that harnesses a neural network model as a regression model that receives a series of noisy accelerometer and gyroscope data and estimates the noise-free velocity values in 2-D space. Taking the integration from the output will provide the position. Three different machine learning models have been used as regression models and their performance has been reported. Three different datasets have been used to evaluate the performance of the proposed method so that sensor values are being collected using integrated IMU sensors on cellphones and the ground truth has been obtained using the cellphone's or external camera\cite{ronin,ridi,oxiod}.

\subsection{Regression Networks}\label{sec:proposed_networks}

The performance of all the data-driven methods as well as the method introduced in \cite{ronin}, highly depends on the regression model and its capacity to do correction, noise cancellation, sensor calibration, and even navigation equations by its architecture and parameters. More importantly, for the end-to-end solutions, these methods are required to perform well in a real-time manner and the inference time, model size, and energy consumption of the models play an important role. For instance, for doing inertial navigation purposes using the cellphone device and only IMU sensors, the less model size, and shorter inference time would be more beneficial in mobile implementation and more applicable for real-time implementation. IMU sensors can be run in the phone for more than 24 hours without having the battery draining problem, and more appropriate models with less size and inference time will help to save even more energy. 

The performance of the mentioned data-driven method with some different state-of-the-art CNN models that have been designed specifically for edge device implementation as its brain has been investigated. Mobilenet\cite{mobilenets}, MobilenetV2\cite{mobilenetv2}, MNAsNet\cite{mnasnet} and EfficientNetB0\cite{efficientnet} have been investigated in this paper. Mobilenet and MobilenetV2 are CNN models that have been designed manually to be implemented on IoT devices by experts. MNAsNet and EfficientNetB0 are edge devices implementing oriented architectures that have been found using a technique called Neural Architecture Search (NAS). In this technique, an architecture for specific purposes (i.e. IoT implementation) is found using a searching method and an optimization algorithm.  Since the input data is 1-D signals, the 1-D version of the components of the networks has been used (1-D Convolution, Batch normalization, and Pooling). 

Other than these models, a new model has been proposed as a regression model for navigation that outperforms these state-of-the-art models in terms of accuracy, inference time, and model size. Also, the testing part of the method introduced in \cite{ronin} has been implemented on an actual edge device and the inference time of all the models on a real device has been reported. For comparison, the mentioned datasets and the datasets that have been introduced in this paper have been used. In the following, the architecture of the state-of-the-art models, as well as the proposed IMUNet model, has been explained.

\subsubsection{Mobilenet\cite{mobilenets}}

Introduced in 2017, Mobilenet is a CNN architecture that has been proposed for mobile and embedded implementation. The architecture harnesses depth-wise and point-wise convolutions instead of regular convolution which reduces computational cost tremendously and makes the architecture appropriate for embedded implementation. The depth-wise and point-wise convolutions are factorized convolutions of the standard convolution. In depth-wise convolution, a single filter is applied to each channel of the input separately and In the point-wise convolution, a 1× 1 convolution is used to merge all the outputs from depth-wise convolutions. However, a standard convolution does the filtering and merging of the result to the new sets of output simultaneously. Since this new architecture separates the filtering and combines the results, the computation cost and model size have been reduced drastically. The figure~\ref{fig:proposed_imunet} demonstrates the factorized elements(depth-wise and point-wise convolutions) of the standard convolution.

\subsubsection{MobilenetV2\cite{mobilenetv2}}
Introduced in \cite{mobilenetv2}, the authors did a little improvement on the structure of the Mobilenet architecture. The idea was to combine the idea of Mobilenet (point-wise and depth-wise convolutions) and use the residual block in a new architecture. In the normal residual block, the residual connection is connecting the bigger channels together. However, in the MobilenetV2, the residual connection is from a narrow or bottleneck channel to the next narrow channel which is called an inverted residual block. For doing the convolutions, instead of using standard convolutions, the idea of Mobilenet has been harnessed to reduce the model size and the computation cost. Moreover, the authors proved that non-linearity will eliminate the information from each inverted residual block to the next one since narrow filters have been connected to each other. Consequently, no nonlinear function has been used at the end of each inverted residual block and the whole block has been called inverted residuals and linear bottleneck.

\subsubsection{MnasNet\cite{mnasnet}}
Designing a CNN model manually for specific purposes is really difficult since balancing the trade-offs between mutually exclusive targets is tedious work when quite a few possible neural architectures can be designed. To address this issue, unlike Mobilenet and MobilenetV2 which have been designed manually, Mnasnet has been found Automatically using a technique called Neural Architecture Search. The authors introduced a method that balances the trade-off between the model's accuracy and its inference latency by executing the inference on a real device. For a search space, a new factorized hierarchical search space has been introduced. This search space allows for the diversity of the different layers in the whole network.
\begin{table*}
    
    \renewcommand\arraystretch{1.6}
      \centering
\begin{tabular}{|*{11}{l|}}
 \hline 
\mcl{\textbf{}}      
    &   \textbf{Test Subjects}            
        &   \textbf{Metric}     &   \textbf{ResNet18}  &   \textbf{MobNet} 
         &   \textbf{MobNetV2}  &   \textbf{MnasNet}  &   \textbf{EffNet} &   \textbf{IMUNet}\\ 
  \hline \hline

  \hline
  
  \mcl{\multirow{2}{*}{Proposed Dataset}}  
        & \multirow{2}{*}{Seen}
            & ATE & \color{blue}2.73 & 2.98 & 3.03 & 2.75 & \color{orange}2.67 & \color{red}2.59  \\ 
            \cline{4-10}\mcl{} 
            &  & RTE & \color{orange}3.03 & 3.42 & 3.55 & \color{blue}3.19 & 3.48 & \color{red}2.97 \\

  \hline \hline
\mcl{\multirow{4}{*}{RONIN Dataset}}  
        & \multirow{2}{*}{Seen}
            & ATE & \color{orange}3.63 & 4.08 & 3.83 & 3.78 & \color{blue}3.66 & \color{red}{3.52}  \\ 
            \cline{4-10}\mcl{} 
            &  & RTE & \color{blue}2.76 & 2.83 & 2.85 &\color{orange} 2.75 & 2.79 & \color{red}2.71 \\
            \cline{3-10}
       \mcl{} & \multirow{2}{*}{Unseen} 
            & ATE & \color{orange}5.65 & 6.16 & 6.17 & \color{red} 5.19 & 5.68 & \color{blue}5.68 \\
            \cline{4-10}\mcl{}  
            &  & RTE & \color{blue}4.57 & 4.75 & 4.69 & \color{orange}4.54 & 4.6 & \color{red}4.49 \\   
  \hline \hline
   \mcl{\multirow{2}{*}{OXIOD Dataset}}  
        & \multirow{2}{*}{Seen}
            & ATE & \color{blue}3.14 & 3.2 & 3.38 & \color{orange}3.08 & 3.22 & \color{red}2.88  \\ 
            \cline{4-10}\mcl{} 
            &  & RTE & \color{blue}2.66 & 2.68 & 2.89 & \color{orange}2.64 & 2.69 & \color{red}2.58 \\
            \cline{3-10}
         
  \hline \hline
\mcl{\multirow{2}{*}{RIDI Dataset}}  
        & \multirow{2}{*}{Seen}
            & ATE & \color{orange}1.56 & 1.73 & \color{red}1.55 & 1.71 & \color{blue}1.67 & \color{orange}1.56  \\ 
            \cline{4-10}\mcl{} 
            &  & RTE & \color{orange}1.92 & 2.09 & \color{blue}1.97 & 2.10 & 2.05 & \color{red}1.83 \\
            \cline{3-10}
       
  \hline \hline

  \mcl{\multirow{1}{*}{PX4 Dataset}}  
        & \multirow{1}{*}{Unseen}
            & ATE & 92.46 & \color{orange}64.94 & 65.86 & \color{red}56.70 & \color{blue}65.42 & 71.66  \\

  \hline 
\end{tabular}
    \caption{Performance evaluation. Five state-of-the-art architectures as well as the proposed architecture have been evaluated on 4 different datasets as well as the proposed datasets. ATE and RTE have been used as a metric in meters. Since the PX4 dataset does not contain the time value, only ATE metric has been calculated for this dataset. The top three results are highlighted in red, orange, and blue colors per row.\\}
\label{tab:model_performance}
\end{table*}

\begin{figure}
  \centering
    \includegraphics[width=3.7in]{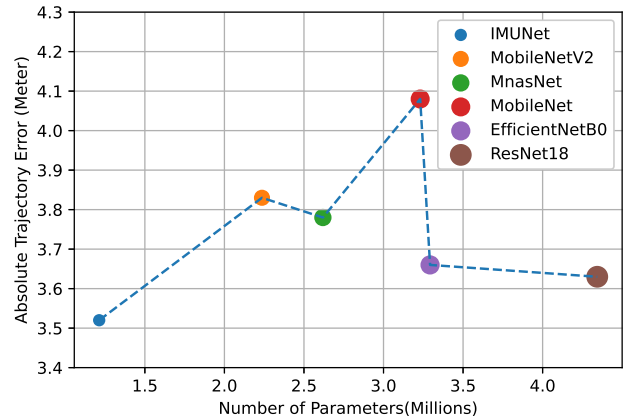}
  \caption{The number of parameters and the accuracy of the networks on the RONIN dataset\cite{ronin}. This figure is similar to the figure~\ref{fig:proposed_flops} with the number of parameters instead of the number of FLOPs}
  \label{fig:proposed_np}
\end{figure}

\subsubsection{EfficientNetB0\cite{efficientnet}}
This model also has been found using the neural architecture search technique. The same technique in \cite{mnasnet} has been used for searching. However, concentrating on a hardware device is not a target since the main idea of the authors is to find a base network and scale that network up to find other architectures. Consequently, a trade-off has been considered between the accuracy and the number of FLOPS. Since this model is the smallest model of this proposed method, we only considered this model from the Efficientnet family as a comparison.

\subsection{IMUNet}
Our experiments show that the performance of all the mentioned models and the ResNet18 model harnessed in \cite{ronin} are close to each other on different datasets. However, the size and inference of those models are different for mobile implementation. ResNet18 is using a regular convolution while all other models are using depth-wise and point-wise convolution instead of regular convolution. Inspiring from Mobilenet\cite{mobilenets}, our experiment shows that replacing the convolutional layers in Resnet18 with depth-wise and point-wise convolutions, not only will it reduce the computational cost of the model drastically, it can improve the accuracy of the model. Since the depth-wise convolution is being run separately on each channel (here different IMU measurements in different dimensions), the noise of each IMU measurement can not affect other IMU outputs during the convolution. By preventing the intervention during the convolution, more purified information can be forwarded to the next layers and the accuracy will improve.

Moreover, the idea of a residual block could pass the input to the output block, transform the pure data, and prevent over-fitting. A new block which is the MobileResNet block has been proposed in this paper. inside the block, depth-wise and point-wise convolution along with a batch normalization has been used to reduce the computational cost and noise intervention between different channels. Residual connection is being used to avoid over-fitting. The rest of the architecture is similar to the Resnet18 introduced in \cite{ronin} which is a one-dimensional version of \cite{resnet}. However,  Exponential Linear Unit(ELU) which is introduced in\cite{elu} has been used as an activation function instead of Rectified Linear Unit(ReLU). It will prevent the dying neurons problem and it will reduce the training time and improve the accuracy. However, according to the equation~\ref{elu}, for negative value, it is slower than ReLU and its derivatives for exponential calculation which increases the latency inference time. However, for the improvement in accuracy, it can be neglected. Figure ~\ref{fig:proposed_imunet} shows the MobileResnet Block architecture. The details of the IMUNet's architecture have been presented in the table~\ref{tab:imunet_table}.
\begin{equation}
\label{elu}
    y=
    \begin{cases}
      x & \text{if}\ x\ge0 \\
      exp(x)-1 & \text{if}\ x < 0
    \end{cases}
  \end{equation}

\begin{figure*}[!t]
\centering
\subfloat[]{\includegraphics[width=3.5in]{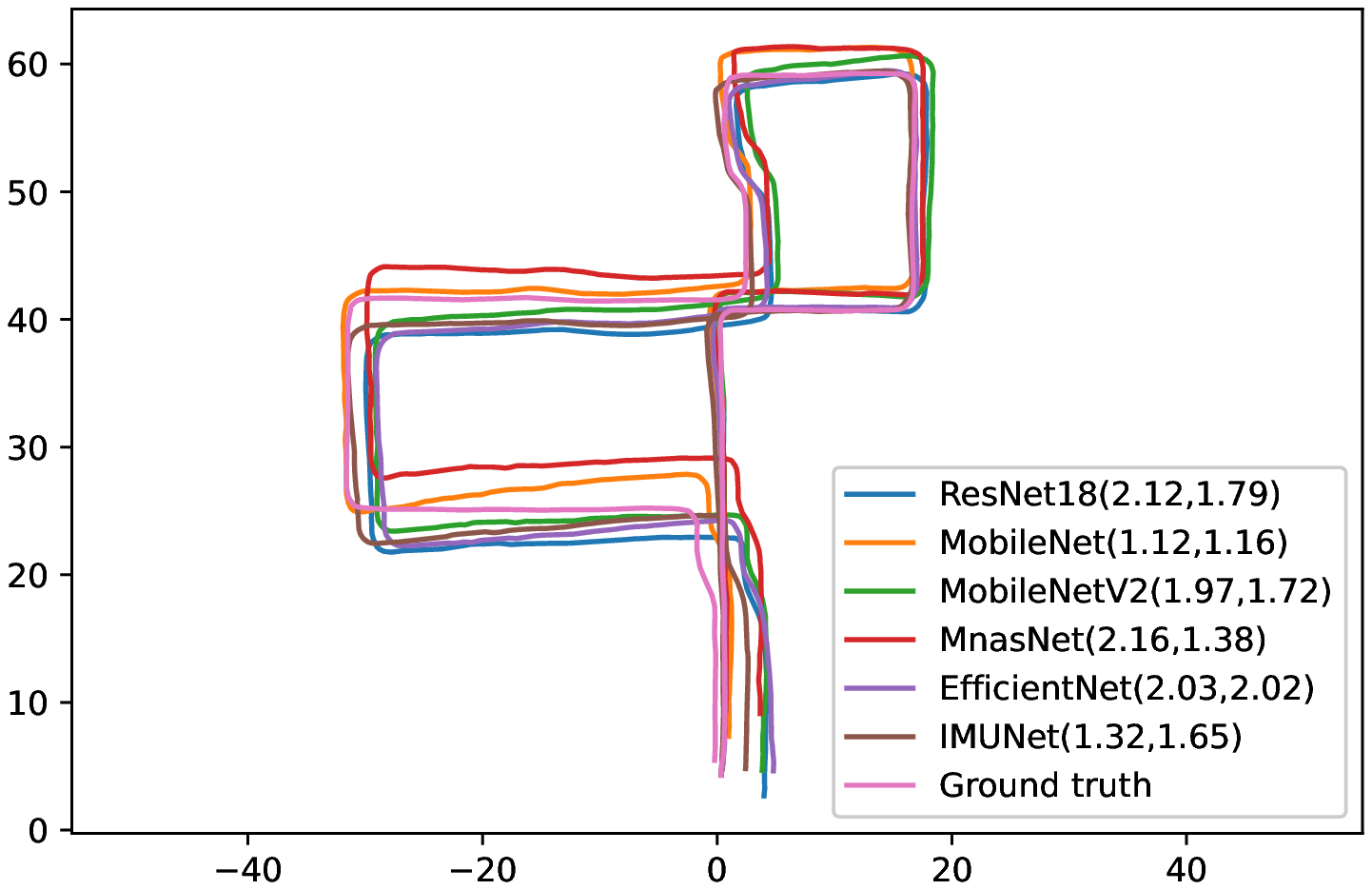}%
\label{fig_first_case}}
\subfloat[]{\includegraphics[width=3.5in]{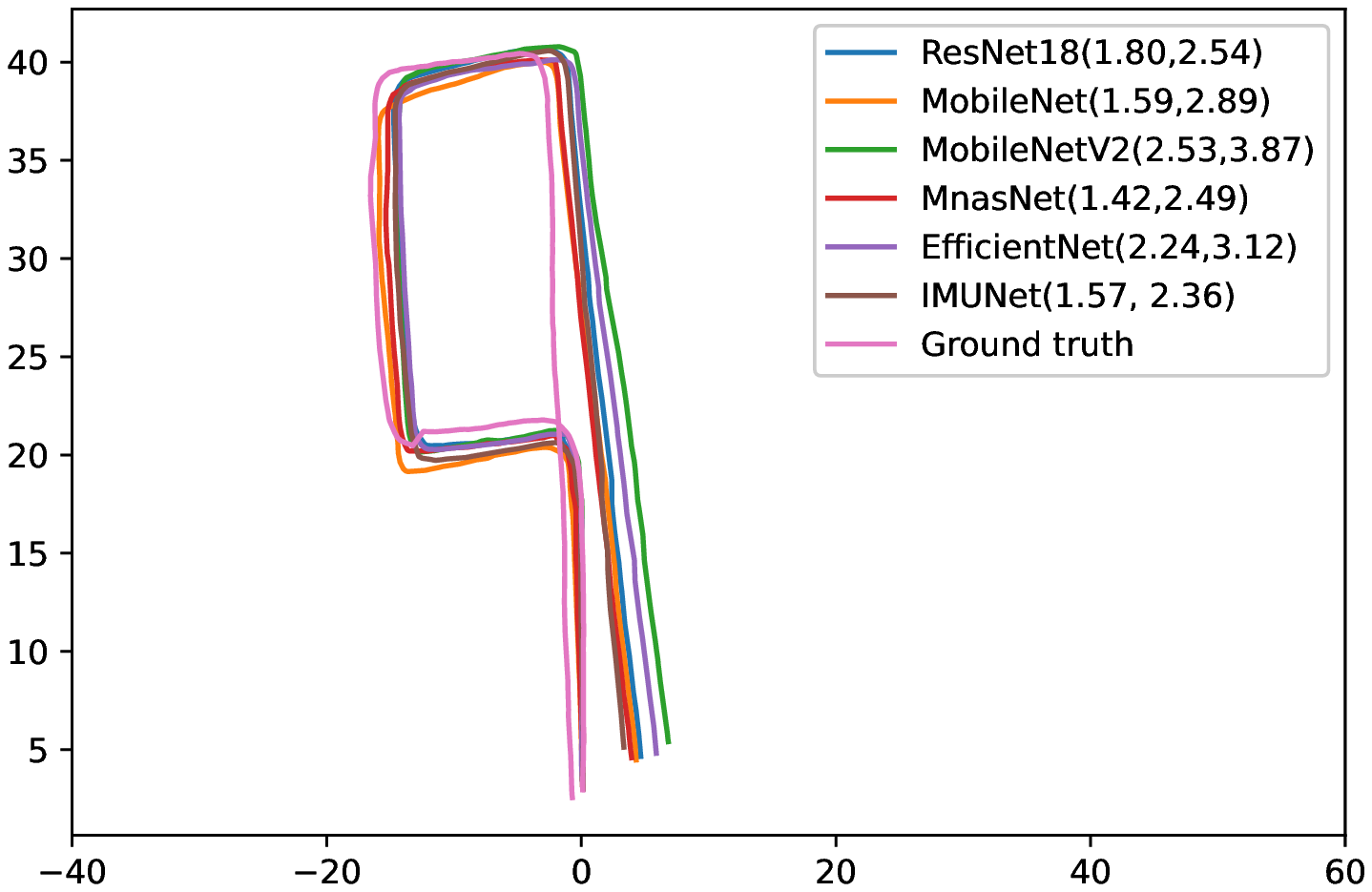}%
\label{fig_proposed_tr}}
\hfil
\centering
\subfloat[]{\includegraphics[width=3.5in]{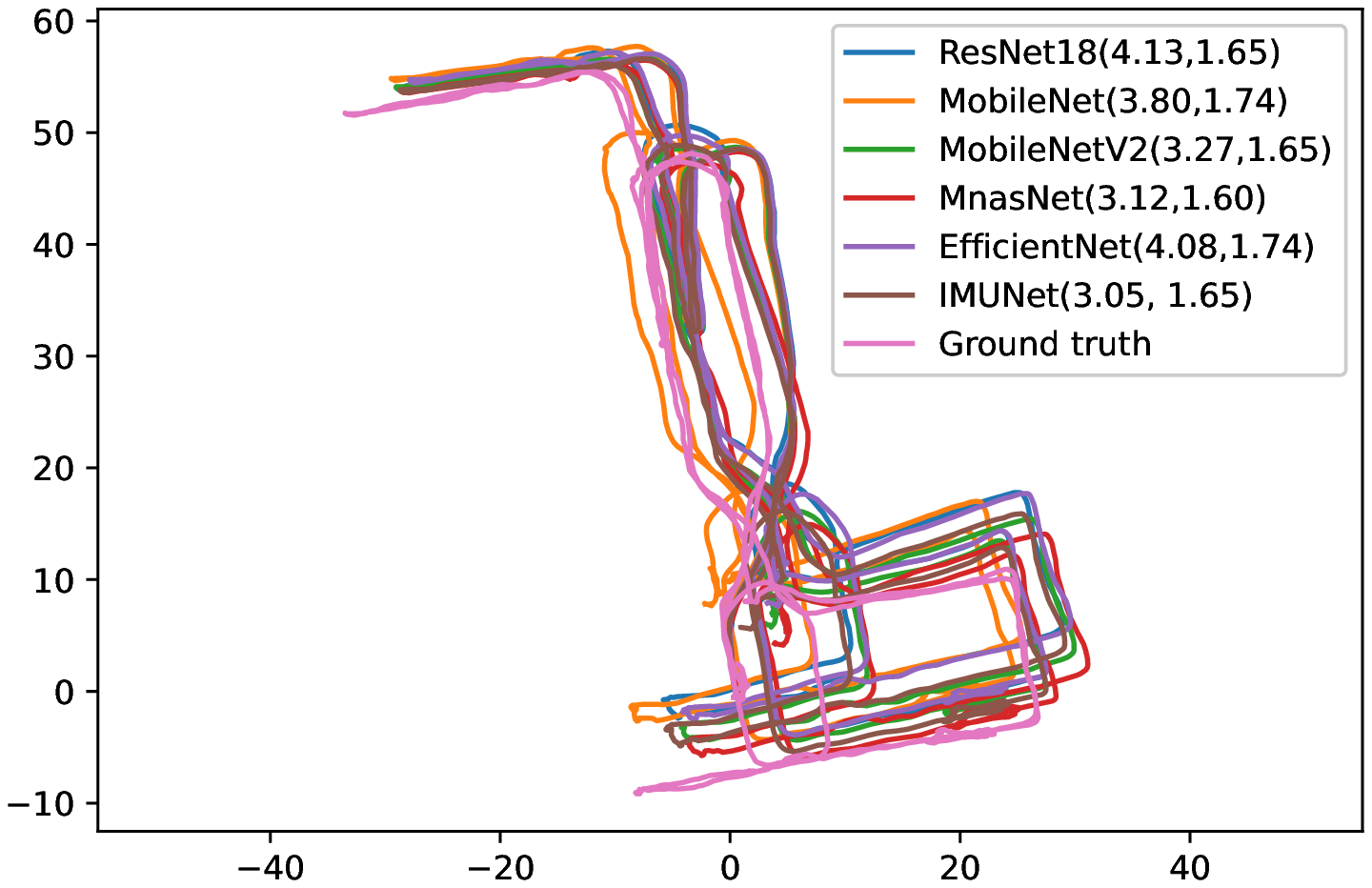}%
\label{fig_first_case_1}}
\subfloat[]{\includegraphics[width=3.5in]{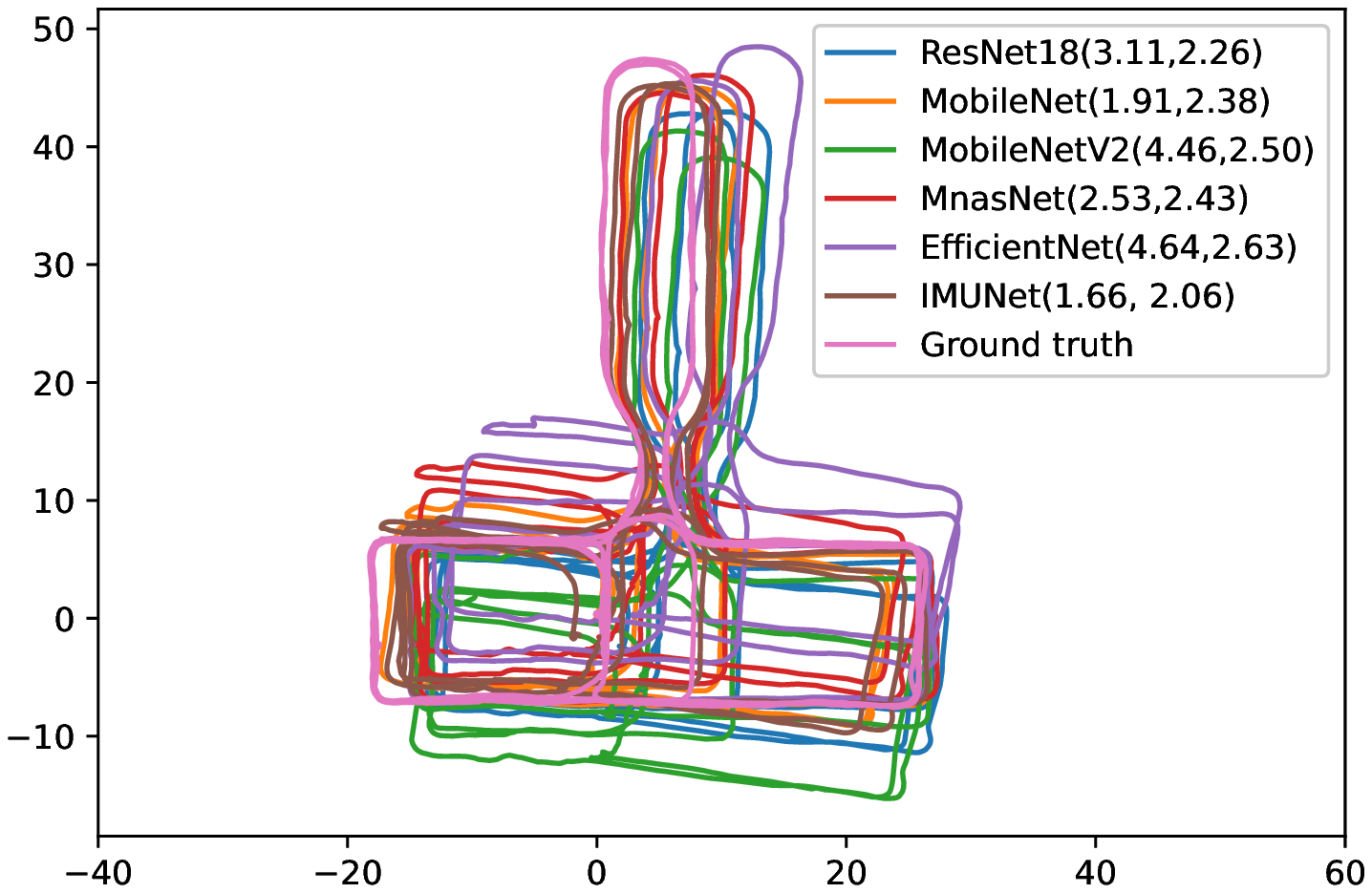}%
\label{fig_ronin_tr}}
\hfil
\subfloat[]{\includegraphics[width=3.5in]{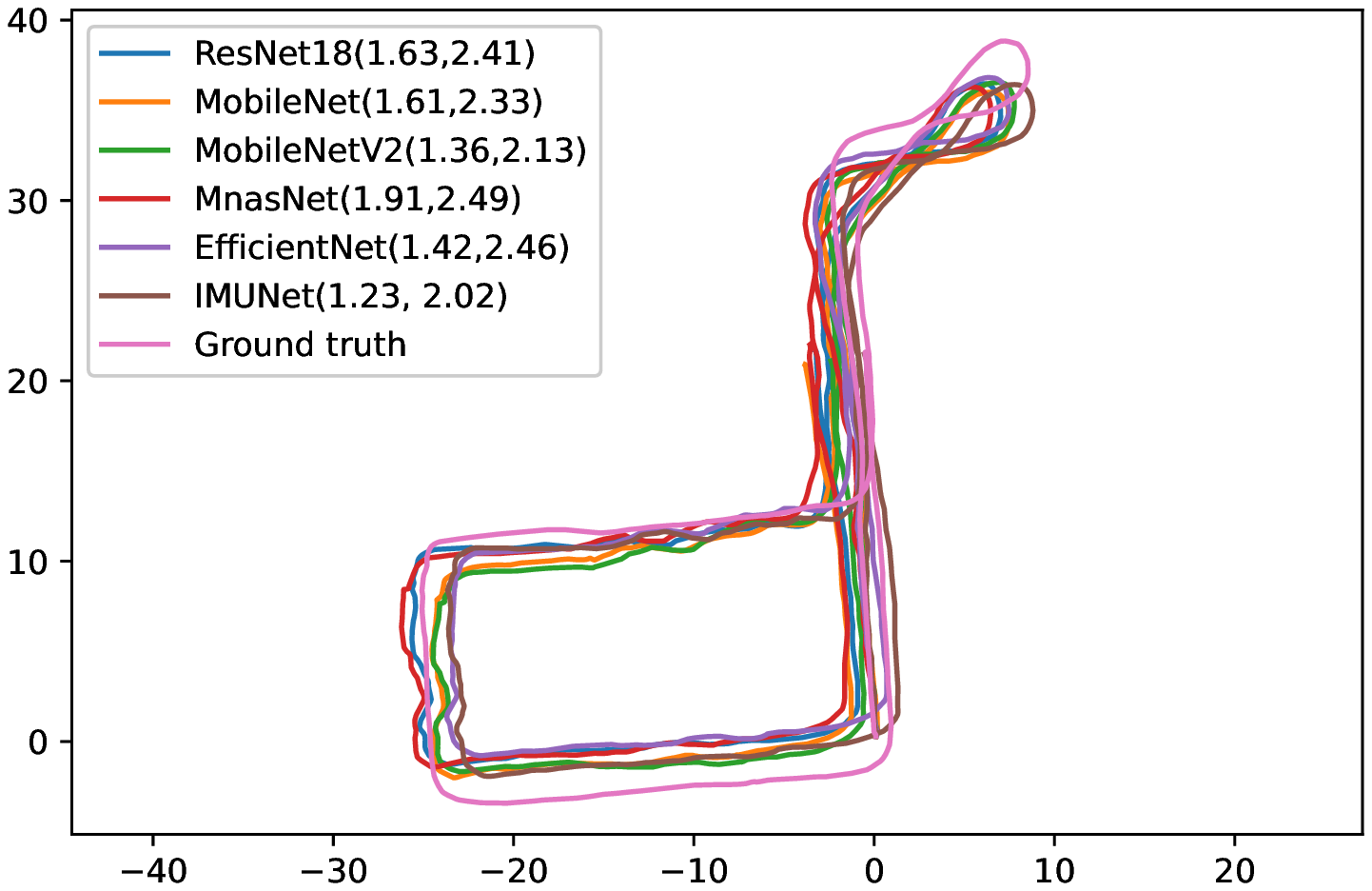}%
\label{fig_first_case_2}}
\subfloat[]{\includegraphics[width=3.5in]{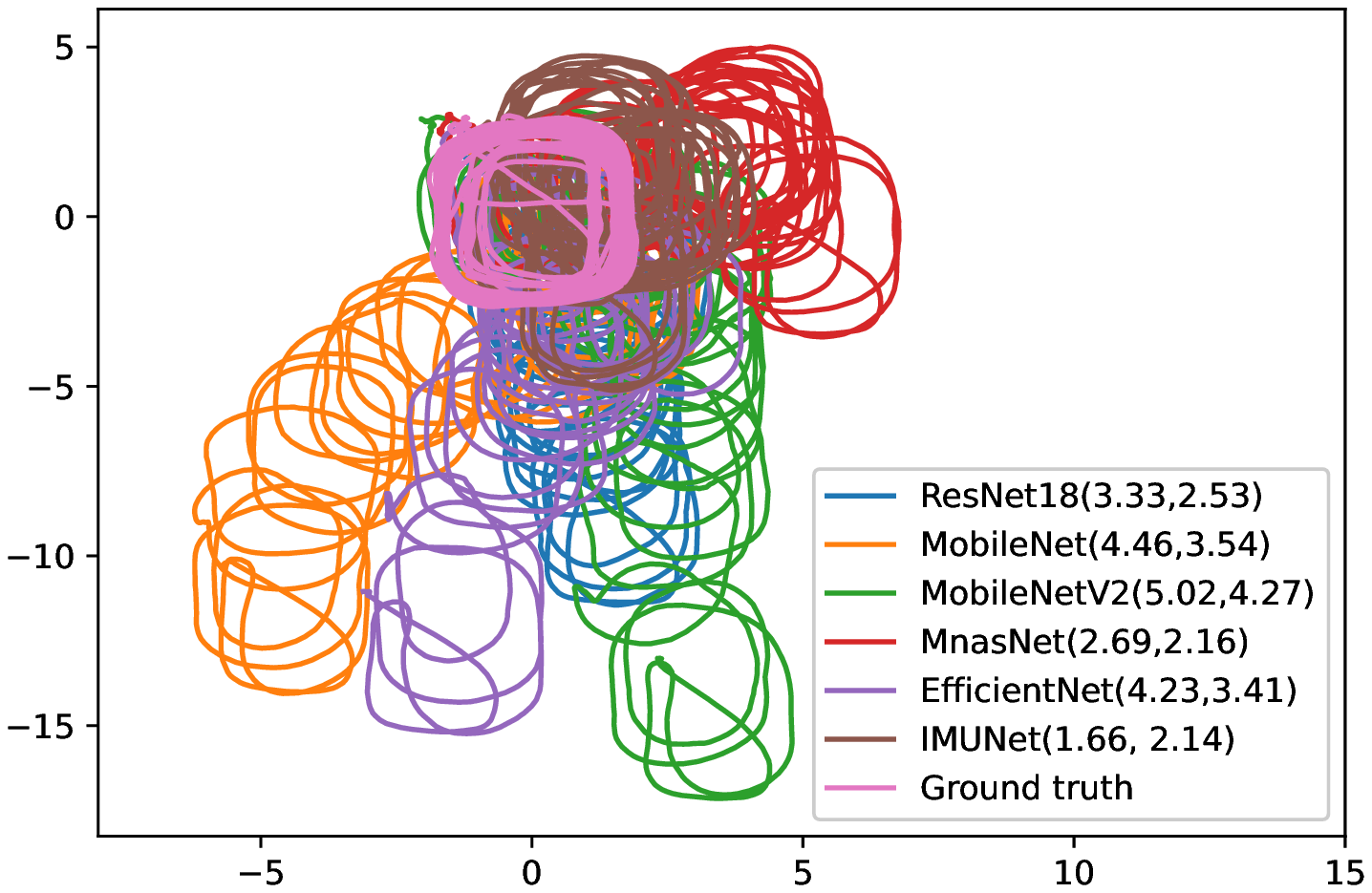}%
\label{fig_ridi_tr}}
\hfil

\caption{Trajectories from the Proposed, RONIN\cite{ronin}, RIDI\cite{ridi}, and OXIOD\cite{oxiod} datasets and the performance of all the state-of-the-art networks. The numbers in the parenthesis show Absolute Trajectory Error (ATE) and the Relative Trajectory Error (RTE) respectively in meters. All the axis are in meters as well. (a) A trajectory from the Proposed dataset with the Tango device. (b) A trajectory from the Proposed dataset using ARCore API for ground truth collection. (c) A trajectory from the RONIN dataset from Seen data (The environment is the same as the training set). (d) A trajectory from the RONIN dataset from Unseen data (The environment is different than the training set). (e) A trajectory from the RIDI dataset. (f) A trajectory from the OXIOD dataset.}
\label{fig_ridi}
\end{figure*}

\begin{figure}[!t]
\centering
\includegraphics[width=3in]{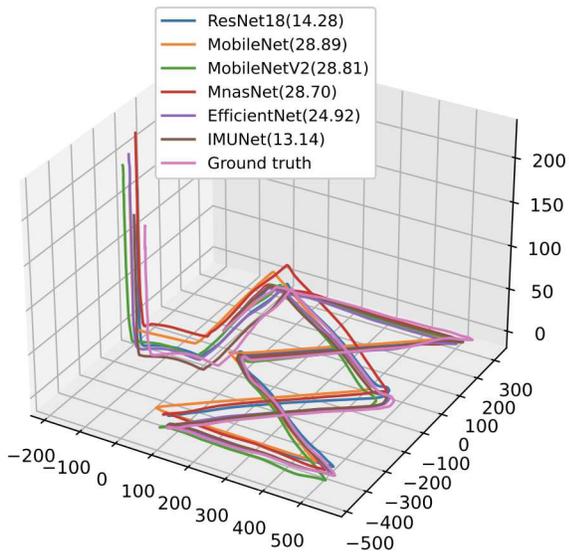}%
\caption{A Trajectory from the PX4 dataset\cite{px4} and the performance of all the state-of-the-art networks (The number in the parenthesis shows Absolute Trajectory Error (ATE) in meters. Axes are in meters as well}
\label{fig_proposed}
\end{figure}

  \begin{table}
    \renewcommand\arraystretch{1.0}
     \resizebox{8.8cm}{!}{
  \begin{tabular}{SSSSS} \\  \toprule
    {$Stage$} & {$Operator$} & {$Resolution$} & {$Channels(n)$} & {$Layer$} \\ \midrule
    1  & {Input} & {6*200} & 6 & {-}\\\midrule
    2  & {Conv1D} & {2*64} & 64 & 1\\\midrule
    3  & MRBlock  &  {2*64}  & 64  & 2  \\
    4  & MRBlock  &  {1*128}  & 128  & 2  \\
    5  & MRBlock  &  {1*256}  & 256  & 2 \\ 
    6  & MRBlock   &  {1*512}  & 512   & 2  \\\midrule
    7  & {Conv1D}  &  {1*128}  & 128  & 1  \\\midrule
    8  & {Dense}  & {1*m} & 512  & 2   \\ \bottomrule
    
\end{tabular}}
\caption{Architecture of IMUNet. m is the number of dimension. }
\label{tab:imunet_table}
\end{table}
\section{Experimental Results}\label{sec:experiment_result}
To evaluate the proposed model, quite a few experiments have been implemented. We evaluate the model on 4 different datasets as well as the proposed dataset. We applied all five datasets to the data-driven method with the mentioned state-of-the-art machine learning models as well as our proposed architecture.

\subsection{Dataset}
For evaluating the performance of the proposed architecture as well as other state-of-the-art models using data-driven methods, quite a few datasets have been harnessed. we applied the methods to all datasets introduced in section ~\ref{sec:available_dataset} that are publicly available. For the RONIN dataset, half of the data is publicly available and is used. all the data in OXIOD and RIDI datasets have been used. For evaluating the models the data in three dimensions with complex maneuvers, some of the px4 dataset logs were downloaded and have been applied to the framework. The proposed dataset which is the combination of using the ARCore and Tango device for ground truth collection has been used to evaluate the architectures as well.
\begin{table*}
    
    \renewcommand\arraystretch{1.6}
      \centering
\begin{tabular}{|*{10}{l|}}
  \hline
\mcl{\textbf{}}      
        &   \textbf{Metric}     &   \textbf{ResNet18}  &   \textbf{MNet} 
         &   \textbf{MNetV2}  &   \textbf{MnasNet}  &   \textbf{EffNet} &   \textbf{IMUNet}\\ 
  \hline 
  \mcl{\multirow{1}{*}{tfLite Size}}  
        
            & Mb & 4.5 & 3.5 & 2.7 & 3.1 & 3.8 & 1.4  \\ 
       \hline    
  \mcl{\multirow{1}{*}{Latency}}  
        
            & $\mu$Sec & 1044 & 907 & 645 & 654 & 967 & 387  \\       
  
  \hline 
\end{tabular}
    \caption{Latency inference on an actual edge device and the tensorflow lite models size of all the state-of-the-art models as well as the proposed model. \\}
\label{tab:real_time_table}
\end{table*}

\subsection{Setup}

Python language has been used for experimental evaluation. We modified the RONIN Resnet method's code provided by the authors by replacing the model with our proposed models and applying the PX4, OXIOD, and the proposed datasets. Pytorch framework has been used for the implementation. Keras's version of the RONIN method was also implemented. However, the results of Pytorch implementation have been reported. Absolute Trajectory Error(ATE) and Relative Trajectory Error (RTE) that were used in \cite{ronin} has been used in all experiments. ATE is the root mean square error(RMSE) and RTE is the average of the ATE in the predefined time interval. A single GPU (Nvidia GeForce GTX 1080 Ti with 11 GB GDDR5X memory) has been used for training all the models. We have used the same parameters as \cite{ronin}. However, each model has been trained for 300 epochs.

\subsection{Model Efficiency}
For evaluating the efficiency of all the models, we considered two common metrics the number of parameters and the number of FLOPs. Although the number of parameters is able to provide a general interpretation of the model efficiency (the less model size means less memory reference and more energy-saving), it does not provide an exact understanding of the number of multiplications during the latency inference. Consequently, both numbers of FLOPs and parameters have been considered to evaluate each model's efficiency. Figure~\ref{fig:proposed_flops} and~\ref{fig:proposed_np}  shows the number of FLOPs and parameters respectively for all the models. The proposed model is outperforming all the state-of-the-art models in terms of accuracy and efficiency.
\subsection{Model performance on Ronin Method}
Table~\ref{tab:model_performance} shows the gist conclusion of the implementation.  The performance of the networks on different datasets has been presented. Out of each dataset, sample trajectories from the test set have been chosen and the performance of all the networks has been depicted in figures~\ref{fig_ridi} and~\ref{fig_proposed}. Unseen trajectories are those subjects that are not included during the training stage. For the PX4 dataset, the subjects are recorded from different people and considered unseen subjects. As it can be seen from the figures and the table, the proposed model outperforms the state-of-the-art edge device-friendly architectures in terms of accuracy.

\subsection{Model Efficiency on Edge devices}
We implemented the test section of the RONIN method with the proposed dataset on the edge devices to evaluate the performance of all models in real-time implementation. For doing so, the TensorFlow lite version of all the models has been created. For Pytorch models, we used the ONNX library which is a third-party library to convert a deep learning model from Pytorch to Keras version. However, for experimental results, we used the models of our Keras implementation. Galaxy S10 cellphone has been used. This application also contains the method using ARCore API for collecting the dataset. Table ~\ref{tab:real_time_table}shows the latency inference and the TensorFlow lite model size of each model as well as the proposed model. The inference time of the proposed model and its size is less than other networks which show its ability to preserve more energy and capability of real-time implementation.

\section{Conclusion}\label{sec:conclusion}
This paper introduces a new architecture for inertial navigation using the sequence of imu measurements. neural inertial navigation methods are the most reliable method to perform navigation and positioning. The performance of these methods highly depends on the capacity of the neural network. More importunately, since the ultimate goal is to perform the navigation on edge devices, the efficiency of the architecture is important for real-time implementation. An accurate and efficient network was proposed in this paper. A new method for collecting a dataset was introduced and the code was shared that allows anyone with a cellphone to collect a dataset for further research. An empirical study was done by implementing and harnessing a one-dimensional version of edge device-friendly state-of-the-art convolutional neural networks for inertial navigation purposes. All the code was shared for modification to enhance further research.

\bibliographystyle{IEEEtran}
\bibliography{IEEEabrv,main_text}

\begin{thebibliography}{10}
\providecommand{\url}[1]{#1}
\csname url@samestyle\endcsname
\providecommand{\newblock}{\relax}
\providecommand{\bibinfo}[2]{#2}
\providecommand{\BIBentrySTDinterwordspacing}{\spaceskip=0pt\relax}
\providecommand{\BIBentryALTinterwordstretchfactor}{4}
\providecommand{\BIBentryALTinterwordspacing}{\spaceskip=\fontdimen2\font plus
\BIBentryALTinterwordstretchfactor\fontdimen3\font minus
  \fontdimen4\font\relax}
\providecommand{\BIBforeignlanguage}[2]{{%
\expandafter\ifx\csname l@#1\endcsname\relax
\typeout{** WARNING: IEEEtran.bst: No hyphenation pattern has been}%
\typeout{** loaded for the language `#1'. Using the pattern for}%
\typeout{** the default language instead.}%
\else
\language=\csname l@#1\endcsname
\fi
#2}}
\providecommand{\BIBdecl}{\relax}
\BIBdecl

\bibitem{li2013high}
M.~Li and A.~I. Mourikis, ``High-precision, consistent ekf-based
  visual-inertial odometry,'' \emph{The International Journal of Robotics
  Research}, vol.~32, no.~6, pp. 690--711, 2013.

\bibitem{9134860}
W.~Liu, D.~Caruso, E.~Ilg, J.~Dong, A.~I. Mourikis, K.~Daniilidis, V.~Kumar,
  and J.~Engel, ``Tlio: Tight learned inertial odometry,'' \emph{IEEE Robotics
  and Automation Letters}, vol.~5, no.~4, pp. 5653--5660, 2020.

\bibitem{zhang2021pose}
M.~Zhang, X.~Zuo, Y.~Chen, Y.~Liu, and M.~Li, ``Pose estimation for ground
  robots: On manifold representation, integration, reparameterization, and
  optimization,'' \emph{IEEE Transactions on Robotics}, vol.~37, no.~4, pp.
  1081--1099, 2021.

\bibitem{nisar2019vimo}
B.~Nisar, P.~Foehn, D.~Falanga, and D.~Scaramuzza, ``Vimo: Simultaneous visual
  inertial model-based odometry and force estimation,'' \emph{IEEE Robotics and
  Automation Letters}, vol.~4, no.~3, pp. 2785--2792, 2019.

\bibitem{levinson2011towards}
J.~Levinson, J.~Askeland, J.~Becker, J.~Dolson, D.~Held, S.~Kammel, J.~Z.
  Kolter, D.~Langer, O.~Pink, V.~Pratt \emph{et~al.}, ``Towards fully
  autonomous driving: Systems and algorithms,'' in \emph{2011 IEEE intelligent
  vehicles symposium (IV)}.\hskip 1em plus 0.5em minus 0.4em\relax IEEE, 2011,
  pp. 163--168.

\bibitem{10.5555/1594745}
J.~Farrell, \emph{Aided Navigation: GPS with High Rate Sensors}, 1st~ed.\hskip
  1em plus 0.5em minus 0.4em\relax USA: McGraw-Hill, Inc., 2008.

\bibitem{Abdulrahim2014UnderstandingTP}
K.~Abdulrahim, T.~Moore, C.~Hide, and C.~Hill, ``Understanding the performance
  of zero velocity updates in mems-based pedestrian navigation,''
  \emph{International Journal of Advancements in Technology}, vol.~5, pp.
  53--60, 2014.

\bibitem{wagstaff2018lstm}
B.~Wagstaff and J.~Kelly, ``Lstm-based zero-velocity detection for robust
  inertial navigation,'' in \emph{2018 International Conference on Indoor
  Positioning and Indoor Navigation (IPIN)}.\hskip 1em plus 0.5em minus
  0.4em\relax IEEE, 2018, pp. 1--8.

\bibitem{HGJHC0_2011_v12n4_371}
\BIBentryALTinterwordspacing
C.~H. Kang, S.~Y. Kim, and C.~G. Park, ``Improvement of a low cost mems
  inertial-gps integrated system using wavelet denoising techniques,''
  \emph{International Journal of Aeronautical and Space Sciences}, vol.~4,
  no.~4, Dec 2011. [Online]. Available:
  \url{http://dx.doi.org/10.5139/IJASS.2011.12.4.371}
\BIBentrySTDinterwordspacing

\bibitem{trawny2005indirect}
N.~Trawny and S.~I. Roumeliotis, ``Indirect kalman filter for 3d attitude
  estimation,'' \emph{University of Minnesota, Dept. of Comp. Sci. \& Eng.,
  Tech. Rep}, vol.~2, p. 2005, 2005.

\bibitem{schneider2017visual}
T.~Schneider, M.~Li, M.~Burri, J.~Nieto, R.~Siegwart, and I.~Gilitschenski,
  ``Visual-inertial self-calibration on informative motion segments,'' in
  \emph{2017 IEEE International Conference on Robotics and Automation
  (ICRA)}.\hskip 1em plus 0.5em minus 0.4em\relax IEEE, 2017, pp. 6487--6494.

\bibitem{chen2018ionet}
C.~Chen, X.~Lu, A.~Markham, and N.~Trigoni, ``Ionet: Learning to cure the curse
  of drift in inertial odometry,'' in \emph{Proceedings of the AAAI Conference
  on Artificial Intelligence}, vol.~32, no.~1, 2018.

\bibitem{jiang2018mems}
C.~Jiang, S.~Chen, Y.~Chen, B.~Zhang, Z.~Feng, H.~Zhou, and Y.~Bo, ``A mems imu
  de-noising method using long short term memory recurrent neural networks
  (lstm-rnn),'' \emph{Sensors}, vol.~18, no.~10, p. 3470, 2018.

\bibitem{ronin}
S.~Herath, H.~Yan, and Y.~Furukawa, ``Ronin: Robust neural inertial navigation
  in the wild: Benchmark, evaluations, \& new methods,'' in \emph{2020 IEEE
  International Conference on Robotics and Automation (ICRA)}.\hskip 1em plus
  0.5em minus 0.4em\relax IEEE, 2020, pp. 3146--3152.

\bibitem{brossard2020denoising}
M.~Brossard, S.~Bonnabel, and A.~Barrau, ``Denoising imu gyroscopes with deep
  learning for open-loop attitude estimation,'' \emph{IEEE Robotics and
  Automation Letters}, vol.~5, no.~3, pp. 4796--4803, 2020.

\bibitem{zhang2021imu}
M.~Zhang, M.~Zhang, Y.~Chen, and M.~Li, ``Imu data processing for inertial
  aided navigation: A recurrent neural network based approach,'' in \emph{2021
  IEEE International Conference on Robotics and Automation (ICRA)}.\hskip 1em
  plus 0.5em minus 0.4em\relax IEEE, 2021, pp. 3992--3998.

\bibitem{cortes2018advio}
S.~Cort{\'e}s, A.~Solin, E.~Rahtu, and J.~Kannala, ``Advio: An authentic
  dataset for visual-inertial odometry,'' in \emph{Proceedings of the European
  Conference on Computer Vision (ECCV)}, 2018, pp. 419--434.

\bibitem{ridi}
H.~Yan, Q.~Shan, and Y.~Furukawa, ``Ridi: Robust imu double integration,'' in
  \emph{Proceedings of the European Conference on Computer Vision (ECCV)},
  September 2018.

\bibitem{arcore}
``{Google}. arcore.'' \url{https://developers.google. com/ar/.}

\bibitem{mobilenets}
A.~G. Howard, M.~Zhu, B.~Chen, D.~Kalenichenko, W.~Wang, T.~Weyand,
  M.~Andreetto, and H.~Adam, ``Mobilenets: Efficient convolutional neural
  networks for mobile vision applications,'' \emph{arXiv preprint
  arXiv:1704.04861}, 2017.

\bibitem{mobilenetv2}
M.~Sandler, A.~Howard, M.~Zhu, A.~Zhmoginov, and L.-C. Chen, ``Mobilenetv2:
  Inverted residuals and linear bottlenecks,'' in \emph{Proceedings of the IEEE
  conference on computer vision and pattern recognition}, 2018, pp. 4510--4520.

\bibitem{mnasnet}
M.~Tan, B.~Chen, R.~Pang, V.~Vasudevan, M.~Sandler, A.~Howard, and Q.~V. Le,
  ``Mnasnet: Platform-aware neural architecture search for mobile,'' in
  \emph{Proceedings of the IEEE Conference on Computer Vision and Pattern
  Recognition}, 2019, pp. 2820--2828.

\bibitem{efficientnet}
M.~Tan and Q.~Le, ``Efficientnet: Rethinking model scaling for convolutional
  neural networks,'' in \emph{International conference on machine
  learning}.\hskip 1em plus 0.5em minus 0.4em\relax PMLR, 2019, pp. 6105--6114.

\bibitem{resnet}
K.~He, X.~Zhang, S.~Ren, and J.~Sun, ``Deep residual learning for image
  recognition,'' in \emph{Proceedings of the IEEE conference on computer vision
  and pattern recognition}, 2016, pp. 770--778.

\bibitem{s130809549}
\BIBentryALTinterwordspacing
A.~G. Quinchia, G.~Falco, E.~Falletti, F.~Dovis, and C.~Ferrer, ``A comparison
  between different error modeling of mems applied to gps/ins integrated
  systems,'' \emph{Sensors}, vol.~13, no.~8, pp. 9549--9588, 2013. [Online].
  Available: \url{https://www.mdpi.com/1424-8220/13/8/9549}
\BIBentrySTDinterwordspacing

\bibitem{li2014high}
M.~Li, H.~Yu, X.~Zheng, and A.~I. Mourikis, ``High-fidelity sensor modeling and
  self-calibration in vision-aided inertial navigation,'' in \emph{2014 IEEE
  International Conference on Robotics and Automation (ICRA)}.\hskip 1em plus
  0.5em minus 0.4em\relax IEEE, 2014, pp. 409--416.

\bibitem{Yang2020OnlineII}
Y.~Yang, P.~Geneva, X.~Zuo, and G.~Huang, ``Online imu intrinsic calibration:
  Is it necessary?'' in \emph{Robotics: Science and Systems}, 2020.

\bibitem{niu2013fast}
X.~Niu, Y.~Li, H.~Zhang, Q.~Wang, and Y.~Ban, ``Fast thermal calibration of
  low-grade inertial sensors and inertial measurement units,'' \emph{Sensors},
  vol.~13, no.~9, pp. 12\,192--12\,217, 2013.

\bibitem{geneva2018lips}
P.~Geneva, K.~Eckenhoff, Y.~Yang, and G.~Huang, ``Lips: Lidar-inertial 3d plane
  slam,'' in \emph{2018 IEEE/RSJ International Conference on Intelligent Robots
  and Systems (IROS)}.\hskip 1em plus 0.5em minus 0.4em\relax IEEE, 2018, pp.
  123--130.

\bibitem{ahmed2018visual}
A.~Ahmed and S.~Roumeliotis, ``A visual-inertial approach to human gait
  estimation,'' in \emph{2018 IEEE International Conference on Robotics and
  Automation (ICRA)}.\hskip 1em plus 0.5em minus 0.4em\relax IEEE, 2018, pp.
  4614--4621.

\bibitem{6851371}
M.~Kourogi and T.~Kurata, ``A method of pedestrian dead reckoning for
  smartphones using frequency domain analysis on patterns of acceleration and
  angular velocity,'' in \emph{2014 IEEE/ION Position, Location and Navigation
  Symposium - PLANS 2014}, 2014, pp. 164--168.

\bibitem{6696807}
D.~G. Kottas, K.~J. Wu, and S.~I. Roumeliotis, ``Detecting and dealing with
  hovering maneuvers in vision-aided inertial navigation systems,'' in
  \emph{2013 IEEE/RSJ International Conference on Intelligent Robots and
  Systems}, 2013, pp. 3172--3179.

\bibitem{deep_nav}
A.~AbdulMajuid, O.~Mohamady, M.~Draz, and G.~El-bayoumi, ``Gps-denied
  navigation using low-cost inertial sensors and recurrent neural networks,''
  \emph{arXiv preprint arXiv:2109.04861}, 2021.

\bibitem{5980229}
L.~Meier, P.~Tanskanen, F.~Fraundorfer, and M.~Pollefeys, ``Pixhawk: A system
  for autonomous flight using onboard computer vision,'' in \emph{2011 IEEE
  International Conference on Robotics and Automation}, 2011, pp. 2992--2997.

\bibitem{7140074}
L.~Meier, D.~Honegger, and M.~Pollefeys, ``Px4: A node-based multithreaded open
  source robotics framework for deeply embedded platforms,'' in \emph{2015 IEEE
  International Conference on Robotics and Automation (ICRA)}, 2015, pp.
  6235--6240.

\bibitem{paul_riseborough}
\BIBentryALTinterwordspacing
P.~Riseborough, R.~Bapst, L.~Meier, C.~Olsson, S.~B. Purohit, M.~Sauder,
  J.~Oes, georgehines, B.~Tak, D.~Agar, W.~Johnson, M.~Charlebois,
  nickolasrossi, pickledgator, and B.~Küng, ``ecl: v0.9.0 release,'' Jun.
  2016. [Online]. Available: \url{https://doi.org/10.5281/zenodo.55367}
\BIBentrySTDinterwordspacing

\bibitem{GARCIA2020136}
\BIBentryALTinterwordspacing
J.~García, J.~M. Molina, and J.~Trincado, ``Real evaluation for designing
  sensor fusion in uav platforms,'' \emph{Information Fusion}, vol.~63, pp.
  136--152, 2020. [Online]. Available:
  \url{https://www.sciencedirect.com/science/article/pii/S1566253520303043}
\BIBentrySTDinterwordspacing

\bibitem{px4}
``{PX4.Community}, px4 flight logs database.''
  \url{https://review.px4.io/browse}.

\bibitem{project_tango}
``{Google}. project tango.'' \url{hhttps://get.google.com/ tango/. 2}.

\bibitem{slam}
H.~Durrant-Whyte and T.~Bailey, ``Simultaneous localisation and mapping (slam):
  Part i the essential algorithms,'' \emph{IEEE ROBOTICS AND AUTOMATION
  MAGAZINE}, vol.~2, p. 2006, 2006.

\bibitem{oxiod}
C.~Chen, P.~Zhao, C.~X. Lu, W.~Wang, A.~Markham, and N.~Trigoni, ``Oxiod: The
  dataset for deep inertial odometry,'' \emph{arXiv preprint arXiv:1809.07491},
  2018.

\bibitem{elu}
\BIBentryALTinterwordspacing
D.-A. Clevert, T.~Unterthiner, and S.~Hochreiter, ``Fast and accurate deep
  network learning by exponential linear units (elus),'' 2015. [Online].
  Available: \url{https://arxiv.org/abs/1511.07289}
\BIBentrySTDinterwordspacing

\end{thebibliography}

\newpage
\section{Biography}
 
\vspace{2pt}

\begin{IEEEbiography}[{\includegraphics[width=1in,height=1.25in,clip,keepaspectratio]{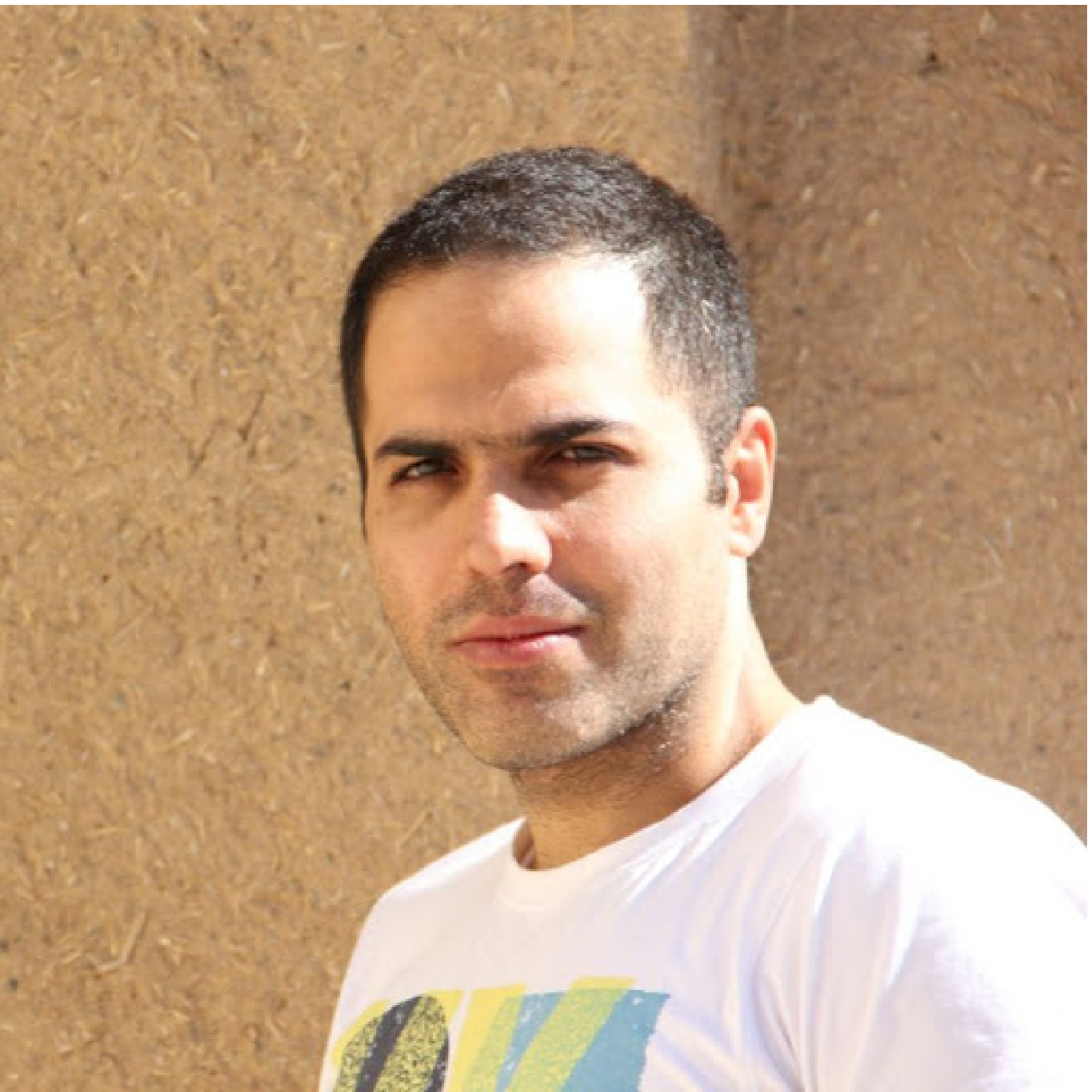}}]{Behnam Zeinali} received his MSc in Electrical Engineering from the Iran University of Science and Technology, Iran, in 2013. From 2013 to 2019 he worked in the industry as a programmer, researcher, and developer in the field of AI. Currently, he is working towards a Ph.D. degree from the University of South Florida. His research focuses on machine and deep learning, computer vision, and mobile application programming.
\end{IEEEbiography}

\vspace{3pt}

\begin{IEEEbiography}[{\includegraphics[width=1in,height=1.25in,clip,keepaspectratio]{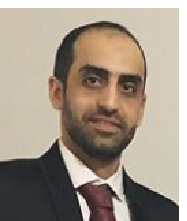}}]{Hadi Zanddizari} is a Machine Learning and Artificial Intelligence Specialist at Ford Motor Company. He received his Ph.D. degree from the University of South Florida in Electrical Engineering. His research interests include deep learning, object detection, semantic segmentation, cybersecurity, and data privacy.
\end{IEEEbiography}
\vspace{3pt}

\begin{IEEEbiography}[{\includegraphics[width=1in,height=1.25in,clip,keepaspectratio]{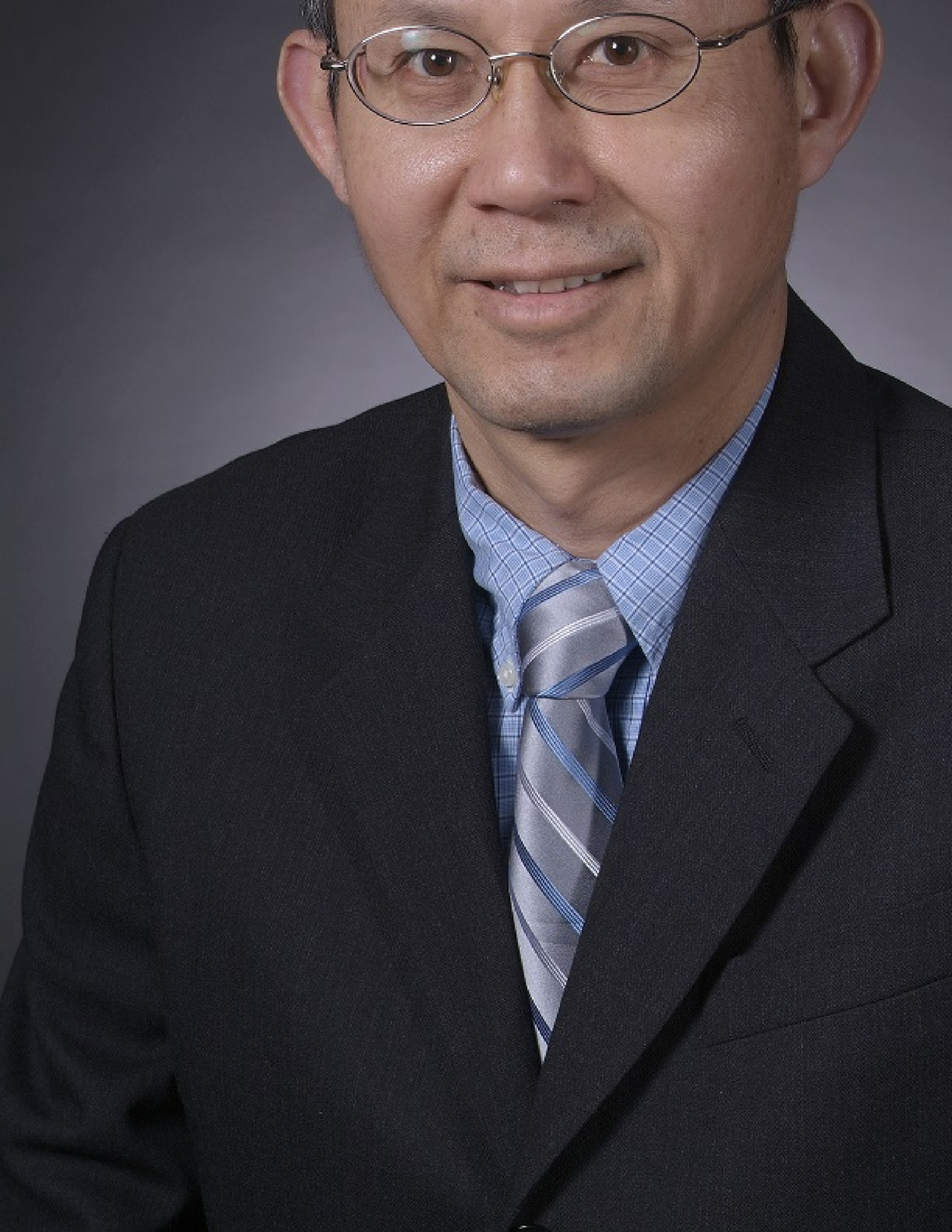}}]{J. Morris Chang} is a professor in the Department of Electrical Engineering at the University of South Florida. He received his Ph.D. degree from the North Carolina State University. His past industrial experiences include positions at Texas Instruments, Microelectronic Center of North Carolina, and AT \& T Bell Labs. He received the University Excellence in Teaching Award at the Illinois Institute of Technology in 1999. He was inducted into the NC State University ECE Alumni Hall of Fame in 2019. His research interests include cyber security and data privacy, machine learning, and mobile computing. He is a handling editor of the Journal of Microprocessors and Microsystems and an editor of IEEE
IT Professional. He is a senior member of IEEE.
\end{IEEEbiography}

\vfill

\end{document}